\documentclass[11pt]{article}

\usepackage[final]{acl}

\usepackage{times}
\usepackage{latexsym}

\usepackage[T1]{fontenc}

\usepackage[utf8]{inputenc}

\usepackage{microtype}

\usepackage{inconsolata}

\usepackage{graphicx}
\usepackage{tabularx}
\usepackage{booktabs}
\usepackage[table]{xcolor} 
\usepackage{soul}
\usepackage{array}
\usepackage{stfloats}
\usepackage{float}
\usepackage{listings}
\usepackage{caption}
\DeclareCaptionType{listing}[Listing][List of Listings]
\usepackage{amsmath}
\usepackage{booktabs}
\usepackage{multirow}
\usepackage{hyperref}
\usepackage{enumitem}

\usepackage{pifont}
\usepackage{pifont}
\usepackage{xcolor}
\usepackage{fontawesome5}
\usepackage{makecell}

\definecolor{colorGreen}{rgb}{0.3, 0.7, 0.3} %
\definecolor{colorMint}{rgb}{0.65, 0.9, 0.65} %
\definecolor{colorPaleGreen}{rgb}{0.4, 0.8, 0.4} %
\definecolor{colorOrange}{rgb}{1.0, 0.7, 0.4} %
\definecolor{colorPink}{rgb}{0.95, 0.7, 0.85} %
\definecolor{colorBlue}{rgb}{0.65, 0.8, 0.95} %

\definecolor{codegreen}{rgb}{0,0.6,0}
\definecolor{codegray}{rgb}{0.5,0.5,0.5}
\definecolor{codepurple}{rgb}{0.58,0,0.82}
\definecolor{backcolour}{rgb}{0.95,0.95,0.92}

\lstdefinestyle{mystyle}{
    backgroundcolor=\color{backcolour},   
    commentstyle=\color{codegreen},
    keywordstyle=\color{magenta},
    numberstyle=\tiny\color{codegray},
    stringstyle=\color{codepurple},
    basicstyle=\ttfamily\footnotesize,
    breakatwhitespace=false,         
    breaklines=true,                 
    captionpos=b,                    
    keepspaces=true,                 
    numbers=left,                    
    numbersep=5pt,                  
    showspaces=false,                
    showstringspaces=false,
    showtabs=false,                  
    tabsize=4,
    extendedchars=true,
    inputencoding=utf8,
    literate={\_}{{\textunderscore}}1
}
\usepackage[many]{tcolorbox}
\tcbuselibrary{listings, skins}
\newtcblisting{codebox}{
  listing engine=listings,
  listing options={
    language=Python,
    basicstyle=\ttfamily\small,
    breaklines=true,
    xleftmargin=2pt
  },
  colback=gray!20,
  colframe=gray!40,
  arc=2mm,
  boxrule=1pt,
  left=2pt,
  right=2pt,
  top=2pt,
  bottom=2pt,
  listing only
}

\newcommand{\hlc}[2]{{\sethlcolor{#1}\hl{#2}}}

\newcommand{\checkmark}{\textcolor{green!60!black}{\ding{51}}}
\newcommand{\xmark}{\textcolor{red!70!black}{\ding{55}}}

\title{\texttt{promptolution}: A Unified, Modular Framework for Prompt Optimization}  %

\author{
 \textbf{Tom Zehle\textsuperscript{1,2,$\ast$}},
 \textbf{Timo Heiß\textsuperscript{3,4,$\ast$}},
 \textbf{Moritz Schlager\textsuperscript{4,5,$\ast$}},\\
 \textbf{Matthias Aßenmacher\textsuperscript{3,4}},
 \textbf{Matthias Feurer\textsuperscript{6,7}}
\\
\textsuperscript{$\ast$}Equal contribution
\textsuperscript{1}ELLIS Institute, Tübingen, Germany
 \textsuperscript{2} University of Freiburg, Germany
 \\
 \textsuperscript{3}LMU Munich, Germany
 \textsuperscript{4}Munich Center for Machine Learning (MCML), Germany
 \\
 \textsuperscript{5}Technical University of Munich, Germany
 \textsuperscript{6}TU Dortmund University, Germany
 \\
 \textsuperscript{7}Lamarr Institute for Machine Learning and Artificial Intelligence, Dortmund, Germany
\\
 \small{
   \textbf{Correspondence:} 
   \href{mailto:tom.zehle@tue.ellis.eu}{tom.zehle@tue.ellis.eu},
   \href{mailto:timo.heiss@stat.uni-muenchen.de}{timo.heiss@stat.uni-muenchen.de},
   \href{mailto:moritz.schlager@tum.de}{moritz.schlager@tum.de}
 }
}

\begin{document}
\maketitle

\begin{figure*}[b!]
    \centering
    \includegraphics[width=.9\textwidth]{figures/promptolution_overview.pdf}
    \caption{Overview of the \texttt{promptolution} framework. \texttt{promptolution} takes a dataset (dev set + few-shot examples), token budget constraints, a description of the task, and optionally initial prompts as input. In an iterative process, a user-selected prompt optimizer refines the prompt(s) by evaluating the LLM's prediction performance on the task's development set and adapting the prompts accordingly (e.g., through another LLM). Detailed logging and callbacks enable tracking the entire process. The optimized prompts are returned and can be evaluated on test data.}
    \label{fig:overview}
\end{figure*}

\begin{abstract}
Prompt optimization has become crucial for enhancing the performance of large language models (LLMs) across a broad range of tasks. Although many research papers demonstrate its effectiveness, practical adoption is hindered because existing implementations are often tied to unmaintained, isolated research codebases or require invasive integration into application frameworks. To address this, we introduce \texttt{promptolution}, a unified, modular open-source framework that provides all components required for prompt optimization within a single extensible system for both practitioners and researchers. It integrates multiple contemporary discrete prompt optimizers, supports systematic and reproducible benchmarking, and returns framework-agnostic prompt strings, enabling seamless integration into existing LLM pipelines while remaining agnostic to the underlying model implementation.

\end{abstract}
\vspace{-.25cm}
\begin{center}
\raisebox{-0.15em}{\faGithub[scale=1.2]}\hspace{0.6em}
\href{https://github.com/automl/promptolution}{AutoML/Promptolution}\\[2pt]
\raisebox{-0.05em}{\hspace{-0.5em}\faYoutube[scale=1.2]}\hspace{0.6em}
\href{https://youtu.be/gySdgjEhsZA}{System Demonstration}
\end{center}

\section{Introduction}

Modern large language models (LLMs) exhibit impressive general-purpose capabilities, being able to solve a wide variety of tasks~\citep{radford-openai19a,ouyang-neurips22a,bai-arxiv23a,touvron-arxiv23a}. 
They can be adapted to solve specific tasks through in-context learning, i.e., simply by a textual instruction and optionally few-shot examples provided to the LLM as input~\citep{brown-neurips20a}. Since this input (referred to as \textit{prompt}) steers the output of the LLM~\citep{karmaker-emnlpfin23a,white-plop23a}, the LLM's performance on a given task highly depends on it -- in terms of quality, formulation, and the choice and order of examples~\citep{zhao-icml21b,lu-aclam22a,zhou-iclr23a}. Table~\ref{tab:prompt-examples} illustrates this with two similar prompts for the GMS8K math dataset~\citep{cobbe-arxiv21a}: despite their strong semantic similarity (matching parts in the same color), their performances differ substantially.
This sensitivity highlights the potential of optimizing prompts for specific tasks, much like how hyperparameter tuning boosts performance in classical machine learning~\citep{kohavi-icml95}. Just as manual hyperparameter search, crafting and refining good prompts by hand (manual prompt engineering) is tedious and unreliable~\citep{jiang-acl20a,liu-acmcs23a}. Similar to how AutoML automates the search over large hyperparameter spaces~\citep{feurer-book19a}, automatic prompt optimization recently emerged to systematically search the combinatorial space of prompts~\citep{li-arxiv25a}.
Especially nowadays, when multi-agent systems have become central in both industry and research, LLMs specialized for individual tasks are increasingly important~\citep{he-acm25a,li-vicinagearth24a}. While fine-tuning entire models is expensive and data-intensive, automatic prompt optimization is a lightweight alternative and potential addition, often compatible with black-box LLMs~\citep{cheng-acl24}.
A diverse collection of prompt optimizers has thus been introduced~\citep[see][]{cui-acl25a,li-arxiv25a}.
However, several hurdles arise when applying these optimizers in practice. Each optimizer lives in its own research repository. Trying and comparing several optimizers requires juggling with multiple code bases and conflicting requirements. Additionally, these repositories are typically not actively maintained, and the research code often lacks proper software tests, documentation, and robustness. Moreover, their setups are inflexible, and deviations from their use cases (often limited to simple classification tasks) and LLM deployments require considerable effort.
While existing libraries and tools for prompt optimization partly address these issues, they are either commercial and closed-source~\citep{amazon,claude,promptperfect,vertex}, only implement a single optimizer~\citep{adalflow,agarwal-arxiv24a,promptimizer,yuksekgonul2024textgrad}, or have a high abstraction level designed mainly for end-to-end AI application development~\citep{khattab-iclr24a,kulin-fruct25a}.

\begin{table}[t]
    \centering
    \footnotesize 
    \begin{tabularx}{0.48\textwidth}{>{\raggedright\arraybackslash} X c}
        \toprule
        \textbf{Prompt} & \textbf{Accuracy} \\
        \midrule
        \hlc{colorPaleGreen}{Tackle}
        \hlc{colorPink}{this elementary math problem}
        \hlc{colorOrange}{by breaking it into logical steps.}
        When you reach the solution,
        \hlc{colorBlue}{enclose the final answer with \texttt{\textless final\_answer\textgreater} and \texttt{\textless/final\_answer\textgreater} markers}
        \hlc{colorMint}{for clarity.}
        & 37.6\% \\
        
        \midrule
        
        \hlc{colorPaleGreen}{Assist with solving}
        \hlc{colorPink}{the elementary or grade school level math problem}
        \hlc{colorOrange}{that requires multiple steps}
        and 
        \hlc{colorBlue}{provide the solution within \texttt{\textless final\_answer\textgreater} \texttt{\textless/final\_answer\textgreater} tags}
        \hlc{colorMint}{for easy identification.}
        & 53.8\% \\
        \bottomrule
    \end{tabularx}
    \caption{Example prompts and their test set accuracy on GSM8K with Llama-3.3-70B (colors = similar phrases).}
    \label{tab:prompt-examples}
\end{table}

\paragraph{Contribution.} We introduce \textbf{\texttt{promptolution}}, a modular, lightweight, and extensible open-source framework for automatic prompt optimization in Python, providing thoroughly tested and stable implementations. Our library implements multiple LLM interfaces, NLP tasks, and contemporary discrete prompt optimization methods, which can be used interchangeably or replaced with custom implementations of the components.
While \texttt{promptolution} facilitates single-prompt optimization for practitioners, a low abstraction level and helpers for systematic, reproducible experiments also make it geared toward researchers. Figure~\ref{fig:overview} shows an overview of how the framework operates.

\paragraph{Outline.} We position our work within the landscape of automatic prompt optimization tools and libraries (\S\ref{sec:related-works}), describe the design of our framework (\S\ref{sec:system-design}), evaluate its performance compared to unoptimized prompts and other libraries to demonstrate its utility and competitiveness (\S\ref{sec:evaluation}), provide use cases and anti-use cases (\S\ref{sec:usage}), and outline future directions of the framework (\S\ref{sec:conclusion}).

\section{Background \& Related Works}\label{sec:related-works}

\paragraph{Prompt Optimization.}

Automatic prompt optimization refers to systematically exploring prompt spaces using various automated optimization strategies, which may optimize both the instruction and the few-shot example components of a prompt~\citep{cui-acl25a,li-arxiv25a,wan-neurips24a}. Prompt optimization is commonly categorized by the nature of the prompt space into continuous~\citep{li-aclam21a,lester-emnlp21a,qin-acl21a} and discrete approaches~\citep{guo-iclr24a,yang-iclr24a,zhou-iclr23a}. For overviews of the field, we refer to \citet{li-arxiv25a} and \citet{cui-acl25a}.
\texttt{promptolution} focuses on the LLM-agnostic and interpretable discrete prompt optimization, which iteratively refines textual prompts directly, often using another ``meta''-LLM\footnote{Can be the same as the one we optimize prompts for.} to generate improved candidates. The following optimizers are implemented in \texttt{promptolution}:

\begin{enumerate}[noitemsep,leftmargin=*,topsep=0pt]
    \item \textbf{\textit{OPRO}}~\citep{yang-iclr24a} uses LLMs as optimizers by providing a task description, examples, and previously scored candidates to the meta-LLM, which then proposes refined instructions.

    \item \textbf{\textit{EvoPrompt}}~\citep{guo-iclr24a} optimizes instructions using evolutionary algorithms, based on (a) a genetic algorithm (GA) and (b) on differential evolution (DE). In both cases, the meta-LLM performs crossover and mutation.

    \item \textbf{\textit{CAPO}}~\citep{zehle-automl25a} is a recent GA-based alternative that leverages AutoML techniques to improve cost-efficiency and jointly optimizes both instructions and few-shot examples, outperforming the discrete optimizers above.
\end{enumerate}

\noindent Other promising prompt optimizers not yet implemented in \texttt{promptolution} include \textit{GEPA}~\citep{agrawal-arxiv25a}, \textit{TextGrad}~\citep{yuksekgonul-nature25a}, \textit{MIPRO}~\citep{opsahl-ong-emnlp24}, and \textit{PromptWizard}~\citep{agarwal-arxiv24a}.

\paragraph{Existing Libraries, Frameworks \& Tools.} Most optimizers above are implemented in siloed research repositories with hard-coded experimental setups. Oftentimes not actively maintained, lacking software tests and proper documentation, they are inherently difficult to use for both scientific benchmark experiments with other optimizers and practical use cases with specific requirements.
In the prompt optimization landscape, many actively maintained libraries and tools have emerged, including both open- and closed-source solutions.

\noindent \textit{Open-Source.} We classify open-source tools along four axes in Table~\ref{tab:rw}: (1) single/multiple optimizers, (2) extensibility, (3) abstraction level, and (4) invasiveness of integration.\footnote{(1) describes whether the tool implements only a single vs. multiple optimizers, (2) whether it can be easily extended to new NLP tasks, LLMs, and optimizers, (3) if the user has direct control over implementation details, and (4) if it is easy to integrate the optimized prompt into arbitrary existing LLM-pipelines.}
In the following, we focus on their most important delimitation criteria compared to \texttt{promptolution}. For explanations of the full categorization in Table~\ref{tab:rw}, see Appendix~\ref{app:categorization}.

Several libraries only implement a single prompt optimizer, including \texttt{TextGrad} \citep{yuksekgonul2024textgrad} using the method from \citet{yuksekgonul-nature25a}, \texttt{AdalFlow} \citep{adalflow} with \textit{LLM-AutoDiff}~\citep{yin-arxiv25a},  %
Microsoft's \texttt{PromptWizard}~\citep{agarwal-arxiv24a} based on the eponymous optimization algorithm, and \texttt{PromptIM} \citep{promptimizer} with its own iterative optimization strategy. \texttt{prompt-ops}~\citep{promptops} implements two optimizers, but focuses on prompt optimization for Llama models.
In contrast, \texttt{promptolution} offers multiple optimizers that can be used interchangeably (1) for arbitrary LLMs (2).

\texttt{DSPy}~\citep{khattab-iclr24a} is arguably the most popular existing framework in prompt optimization for building modular, declarative LLM pipelines and includes an embedded prompt optimization component. It supports multiple optimizers like \textit{MIPROv2} \citep[building on][]{opsahl-ong-emnlp24} or \textit{GEPA}~\citep{agrawal-arxiv25a}, and can combine LLM training with prompt optimization~\citep{soylu-emnlp24a}. While \texttt{DSPy} integrates prompt optimization as part of a monolithic, high-level program compilation procedure, \texttt{promptolution} exposes optimization as an explicit, iterative process, enabling finer control over optimization dynamics, intermediate results, and budget-aware stopping. Consequently, \texttt{promptolution} focuses exclusively on prompt optimization at a lower level of abstraction (3), making it geared toward researchers for systematic benchmarking and advanced practitioners rather than end-to-end AI application development.
Moreover, \texttt{DSPy} only integrates with LLM applications in the \texttt{DSPy} framework while integration into other implementations requires larger refactoring (4). In contrast, \texttt{promptolution} returns a prompt string, enabling integration by directly replacing the existing prompt with the optimized one in any arbitrary LLM application (details in Appendix~\ref{app:dspy}).
\texttt{promptomatix}~\citep{murthy-arxiv25a} builds on \texttt{DSPy} through its structured prompt compilation backend while also offering a lighter meta-prompt optimizer.

\texttt{CoolPrompt}~\citep{kulin-fruct25a} is an LLM-agnostic framework that supports multiple optimizers and emphasizes ``zero-configuration'' (3) in contrast to \texttt{promptolution}, where researchers maintain close control over the setup.

\begin{table}[t]
    \scriptsize 
    \centering
    \begin{tabular}{l|c|c|c|c}
    \toprule
    \multirow{2}{*}{\textbf{Framework}} & Multiple & Extensible & Low & Non-Invasive\\
     & Optimizers &  & Abstraction & Integration\\
    \midrule
    \textbf{\texttt{promptolution}} & \checkmark & \checkmark & \checkmark & \checkmark \\
    DSPy & \checkmark & \checkmark & \xmark & \xmark \\
    CoolPrompt & \checkmark & (\checkmark) & \xmark & \checkmark \\ 
    promptomatix & \checkmark & \xmark & \xmark & \xmark\\
    prompt-ops & \checkmark & \xmark & \checkmark & \checkmark \\
    TextGrad & \xmark & \xmark & \checkmark & \xmark \\
    AdalFlow & \xmark & \xmark & \checkmark & \xmark \\
    PromptWizard & \xmark & \xmark & \checkmark & \checkmark \\
    PromptIM & \xmark & \xmark & \checkmark & \checkmark \\
    \bottomrule
    \end{tabular}
    \caption{Comparison of open-source frameworks.}
    \label{tab:rw}
\end{table}

Other related tools with a slightly different focus include \texttt{PromptBench}~\citep{zhu-jmlr24a}, which targets LLM evaluation supporting only simple optimization techniques, and \texttt{OpenPrompt}~\citep{ding-aclam22a}, designed for prompt learning for language models predating modern LLMs. %

\noindent\textit{Closed-Source.} Proprietary tools range from commercial web-platforms like \texttt{PromptPerfect}~\citep{promptperfect}, cloud integrations such as Google Cloud’s \textit{Vertex AI Prompt Optimizer}~\citep{vertex} and the \textit{AWS Bedrock Prompt Engineering Playground}~\citep{amazon}, to vendor-specific solutions like \textit{Anthropic Claude Prompt Tools}~\citep{claude}.

\paragraph{Positioning of \texttt{promptolution}.} Our library is an open-source, LLM-agnostic, and highly modular framework. It focuses exclusively on prompt optimization rather than constructing full LLM pipelines. It already includes relevant LLM interfaces and NLP tasks, along with evaluation metrics, and provides multiple contemporary discrete prompt optimizers in a single, unified system. The framework is highly customizable and extensible, offering fine-grained control over optimizers, tasks, logging, evaluation, and experiment configuration, and can be seamlessly integrated into existing LLM applications. As a result, it is suitable for both practitioners performing single-prompt optimization and for researchers conducting systematic, reproducible large-scale benchmark studies.

\section{System Design}\label{sec:system-design}

\begin{figure}[b]
    \centering
    \includegraphics[width=0.9\linewidth]{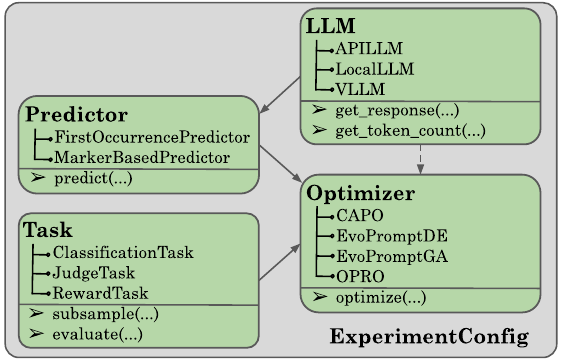}
    \caption{Core components of \texttt{promptolution}. The upper part of each box lists the component implementations, the lower part important functions. Arrows between components indicate conceptual connections.} 
    \label{fig:classes}
\end{figure}

\texttt{promptolution} is designed as a modular and extensible framework consisting of four key components (see Figure~\ref{fig:classes}). All components follow a unified interface defined through corresponding \texttt{Base}-classes, ensuring that implementations can be used interchangeably, remain fully compatible with the framework, and automatically inherit shared functionality. While each component can be configured individually (see \S\ref{sec:llm}--\ref{sec:optimizer}), a separate \texttt{ExperimentConfig} together with associated helper functions enables convenient parameterization of all components in a single object (see \S\ref{sec:experiment-config}).

\subsection{LLM}\label{sec:llm}

The \texttt{LLM} component provides an interface for obtaining responses from any LLM implementation. Its base class enables parallelization and monitors token usage. Three classes are implemented:

\begin{enumerate}[noitemsep,leftmargin=*,topsep=0pt]
    \item \textbf{\texttt{APILLM}} enables calls to LLMs hosted via an API, covering common vendors such as OpenAI and Anthropic. Listing~\ref{listing:manual-setup}~\hyperref[listing:manual-setup]{(L.1--5)} illustrates the setup through the DeepInfra API. 
    \item \textbf{\texttt{LocalLLM}} allows using a local model via the \texttt{transformers} library~\citep{wolf-emnlpsys20a}.
    \item \textbf{\texttt{VLLM}} integrates the \texttt{vllm} library~\citep{kwon-sosp23a} for efficient high-throughput inference and serving, and deployment on GPU clusters.
\end{enumerate}

\subsection{Predictor}\label{sec:predictor}

The \texttt{Predictor} component defines how predictions are extracted from LLM output. While the \texttt{FirstOccurrencePredictor} searches and extracts  the first occurrence of any possible class label in the response, the more robust \texttt{Marker-} \texttt{BasedPredictor} (Listing~\ref{listing:manual-setup},~\hyperref[listing:manual-setup]{L.6}) extracts between predefined HTML-like markers (see Table~\ref{tab:marker-extraction}).

\begin{table}[t]
    \centering
    \footnotesize 
    \begin{tabularx}{0.48\textwidth}{X} %
        \\
        \toprule
        To solve this problem, we need to calculate the total number of fish in the fishbowls at all the tables. First, [...] Then, we add the 3 fish from the table that has 3 fish: 62 fish + 3 fish = 65 fish.
        \hlc{colorBlue}{\texttt{\textless final\_answer\textgreater}}
        \hlc{colorPink}{65}
        \hlc{colorBlue}{\texttt{\textless /final\_answer\textgreater}}      \\
        \bottomrule
    \end{tabularx}
    \caption{Example of the \texttt{MarkerBasedPredictor} extraction from a LLM response for a GSM8K sample.}
    \label{tab:marker-extraction}
\end{table}

\subsection{Task}\label{sec:task}

The \texttt{Task} component holds the dataset and other task-related information, including a textual description of the task, and defines how prompts are evaluated. It controls how subsampling is performed, which is crucial for efficiency, as not every prompt needs evaluation on the full data for a reasonable performance estimate.\footnote{Depending on the subsampling strategy, evaluation is either performed on randomly drawn samples, a slice of the dataset (block), or the full dataset.}
We implement the following tasks relevant to NLP:

\begin{enumerate}[noitemsep,leftmargin=*,topsep=0pt]
    \item \textbf{\texttt{ClassificationTask}} targets discrete class labels and evaluates predictions using standard classification metrics. In our example in Listing~\ref{listing:manual-setup}~\hyperref[listing:manual-setup]{(L.~7--13)}, the task is specified using accuracy as metric. The task is easily created from a \texttt{Pandas DataFrame} by specifying the input and label columns, along with a task description.
    
    \item \textbf{\texttt{JudgeTask}} is based on LLM-as-a-judge~\citep{zheng-neurips23a}, where another LLM scores the quality of the output according to the task description. This enables optimizing for subjective, creative tasks, both supervised and unsupervised, as ground-truth labels are optional.
    
    \item \textbf{\texttt{RewardTask}} allows optimization w.r.t. custom reward functions. The reward can represent any measurable objective, such as code execution time or business metrics. Users define their own reward functions that compute a score (reward) directly from a predictor’s output.
\end{enumerate}

\subsection{Optimizer}\label{sec:optimizer}

The \texttt{Optimizer} component combines all other components by using a \texttt{Predictor} and a \texttt{Task} to determine the optimal prompt for the specified setup. Depending on the chosen optimization method, it may also rely on a (meta-) \texttt{LLM} for prompt alteration. As the core of the framework, it iteratively evaluates LLM predictions for a given task and refines the prompt(s) according to the respective optimization strategy.

\texttt{promptolution} currently implements four established prompt optimizers: \texttt{OPRO}~\citep{yang-iclr24a}, both \texttt{EvoPromptDE} and \texttt{EvoPromptGA}~\citep{guo-iclr24a}, and the current SOTA \mbox{discrete} prompt optimizer \texttt{CAPO}~\citep{zehle-automl25a}. \texttt{promptolution} caches previously evaluated prompts out of the box, and constrains prompt evaluation during optimization to a subset of the available data, making algorithms faster and more efficient. In Listing~\ref{listing:manual-setup}, we set up \texttt{CAPO}~\hyperref[listing:manual-setup]{(L.~14--22)} and let it optimize prompts for 12 steps~\hyperref[listing:manual-setup]{(L.~23)}.%

Furthermore, new prompt optimizers can be added with minimal effort by inheriting from the base optimizer class and implementing a custom \texttt{\_step()} method that defines the iterative optimization scheme. This makes our framework useful for researchers developing and benchmarking new optimization algorithms.

\begin{figure}[h]
\footnotesize
\begin{lstlisting}[language=Python,basicstyle=\ttfamily\small,xleftmargin=20pt,numbers=left,breaklines=true]
llm = APILLM(
    api_url="api.deepinfra.com/v1/...",
    model_id="google/gemma-3-27b-it",
    api_key="...",
)
predictor = MarkerBasedPredictor(llm=llm)
task = ClassificationTask(
    df,
    task_description="The task is...",
    x_column="text",
    y_column="label_text",
    metric=accuracy_score,
)
optim = CAPO(
    predictor,
    task,
    meta_llm=llm,
    init_prompts=[
        "Classify the text based on... ",
        # ...
    ],
)
prompts = optim.optimize(n_steps=12)
\end{lstlisting}
\caption{Setup of \texttt{promptolution}'s core components.}
\label{listing:manual-setup}
\end{figure}

\subsection{Experiment Configuration \& Helper}\label{sec:experiment-config}

\texttt{promptolution} not only supports optimizing prompts for a single specific setup but also provides an \texttt{ExperimentConfig} framework that enables a convenient, structured configuration for larger benchmark experiments. Additional helper functions enable the running and evaluation of such experiments with just a few lines of code.

Listing~\ref{listing:config-setup}~\hyperref[listing:config-setup]{(L.~1--7)} illustrates how the previous example (Listing~\ref{listing:manual-setup}) can be expressed with a single config class. Arguments that are not specified resort to carefully chosen defaults, while arguments that were set but not used during initialization of the classes will throw a warning. The associated helper functions allow users to run the optimization process according to the config (\texttt{run\_optimization}) and to evaluate the prompts on unseen test data (\texttt{run\_evaluation}), without requiring manual initialization of the various classes affected. The \texttt{run\_experiment} function (Listing~\ref{listing:config-setup},~\hyperref[listing:config-setup]{L.~8}) combines both steps, to support researchers who want to perform extensive benchmark studies across multiple datasets and optimizers.

\begin{figure}[h]
\footnotesize
\begin{lstlisting}[language=Python,basicstyle=\ttfamily\small,xleftmargin=20pt,numbers=left,breaklines=true]
config = ExperimentConfig(
    optimizer="capo",
    task_description="The task is...",
    n_steps=12,
    api_url="api.deepinfra.com/v1/...",
    model_id="google/gemma-3-27b-it",
)
prompts = run_experiment(df, config)
\end{lstlisting}
\caption{Running an experiment via the \texttt{ExperimentConfig} abstraction.}
\label{listing:config-setup}
\end{figure}

\subsection{Supporting Modules and Utilities}\label{sec:utils} %

\textit{Exemplar Selection}: Some prompt optimization algorithms (e.g., OPRO or EvoPrompt) do not consider few-shot examples. However, they can substantially improve LLM performance~\citep{brown-neurips20a}, even with simple selection strategies~\citep{wan-neurips24a}. \texttt{promptolution} offers post-hoc exemplar selection in an additional module, implementing random selection and random search\footnote{Random selection selects exemplars at random, whereas random search generates multiple sets of random examples, evaluates them, and selects the best performing set.} to add few-shot examples to a fixed instruction.

\textit{Initial Prompt Creation}: Many prompt optimizers require an initial pool of prompts to start with. We offer functions to automatically create prompts from a task description, a base prompt~\cite[following][]{zhou-iclr23a}, or samples from a dataset.

\textit{Callbacks}: The base class for the optimizers supports callbacks, allowing for easy tracking of the optimization progress. Callbacks can access the state of the optimizer at every optimization step, and optionally terminate the process. Important implementations include the \texttt{TokenCountCallback}, which tracks the accumulated token budget and terminates optimization if a specified threshold is exceeded, and the \texttt{FileOutputCallback}, which writes prompts and their scores to a file, enabling easy post-hoc analysis of the process. %

\section{Evaluation}\label{sec:evaluation}

\paragraph{Setup.} To evaluate \texttt{promptolution} and contextualize its performance relative to other prompt optimization tools, we perform prompt optimization on the popular \textit{GSM8K}~\citep[grade school math word problems;][]{cobbe-arxiv21a} and \textit{SST-5} dataset~\citep[sentiment classification;][]{socher-emnlp13a}. We use \texttt{gemma-3-27B instruction tuned}~\citep{kamath-arxiv25a} as downstream LLM, and, for optimizers that require one, also as meta-LLM. Further details on datasets and implementation choices are provided in Appendix~\ref{app:exp}.

For our comparison, we employ the optimizers \textit{CAPO}, \textit{EvoPromptGA}, and \textit{OPRO} from the \texttt{promptolution} library, and additionally evaluate two other frameworks with leading prompt optimizers: \texttt{AdalFlow} (\textit{LLM-AutoDiff}) and \texttt{DSPy} (\textit{GEPA}). To assess the impact of prompt optimization itself, we also evaluate three unoptimized zero-shot prompts (see Appendix~\ref{app:exp}) for each dataset and report their average performance.
We use the default parameterization for each optimizer to ensure comparability with practical use cases, where users often lack the budget for an extensive hyperparameter search. We further restrict the token budget to at most one million in- and output tokens combined, which corresponds to a cost below \$0.15 for this LLM. All frameworks are initialized from a single task description, without any initial prompts, to test the full automation workflow. While \texttt{DSPy} and \texttt{AdalFlow} use the task description as a starting point for their respective optimizers, \texttt{promptolution}'s optimizers additionally require a set of initial prompts. These are generated via \texttt{promptolution}'s utility function \texttt{create\_prompts\_from\_task\_description}. Both the unoptimized prompts and best prompts per optimizer (based on development-set performance) are evaluated on an unseen test set.

For reproducibility, we make the experiment scripts, seed, and raw results publicly available at \url{https://github.com/finitearth/prompt-optimization-framework-comparison}.

\begin{table}[b]
    \centering
    \footnotesize 
    \begin{tabular}{llcc} 
        \toprule
        \textbf{Framework} & \textbf{Optimizer} & \textbf{GSM8K} & \textbf{SST5} \\
        \midrule
        \textit{Baseline} & \textit{unoptimized} & 78.1 & 44.6 \\ 
        \midrule
        \texttt{AdalFlow} & AutoDiff & 88.7 & 55.7 \\
        \texttt{DSPy} & GEPA & 84.7 & 42.0 \\ 
        \midrule
        \multirow{3}{*}{\texttt{promptolution}} & OPRO & 69.7 & \underline{56.0} \\
         & EvoPrompt & \underline{91.0} & 53.3 \\
         & CAPO & \textbf{93.7} & \textbf{56.3} \\
        \bottomrule
    \end{tabular}
    \caption{Test set accuracy of optimized prompts using Gemma3-27B-it. Bold values indicate the best, underlined values the second-best performance per dataset.}
    \label{tab:eval}
\end{table}

\paragraph{Results.} A summary of the results is presented in Table~\ref{tab:eval}. With the exception of \textit{GEPA} on SST-5 and \textit{OPRO} on GSM8K, all optimizers substantially outperform the unoptimized baseline with improvements of up to 15\%p in accuracy (\texttt{CAPO} on GSM8K). This demonstrates the utility of prompt optimization in general and of our framework in particular. The best-performing optimizer on both datasets, \textit{CAPO}, as well as each runner-up, is implemented in \texttt{promptolution}, underscoring the competitiveness of our framework compared to other prompt optimization tools.
Although not every optimizer included in \texttt{promptolution} performs optimally on every task (e.g., \textit{OPRO} on GSM8K), the library consistently provides at least one strong optimizer per task. Combined with its modular design, this allows users to switch easily to an alternative optimizer if the current one yields unsatisfying performance. We further emphasize that comparing optimizers within \texttt{promptolution} required minimal manual effort due to its dedicated support for systematic benchmark experiments.

For more in-depth evaluations of optimizers utilizing the \texttt{promptolution} framework, we point to \citet{zehle-automl25a}.%

\section{Use Cases \& Anti-Use Cases}
\label{sec:usage}

The choice of prompt optimization frameworks depends not solely on performance, but also on \emph{integration constraints}, the \emph{scope of optimization}, and tolerance for \emph{framework lock-in}. We outline four common use cases to clarify these trade-offs.

\paragraph{Case I: Integration in Existing Pipelines.}
\textit{A practitioner already operates custom LLM pipelines and aims to improve performance solely via better prompts.} \texttt{promptolution} returns a single optimized prompt string that can be directly substituted into existing inference pipelines without refactoring application logic. Since optimized prompts are plain text, users retain full flexibility regarding LLM providers, deployment modes (API, local models, or vLLM), and future model updates. In contrast, adopting frameworks in which prompts are framework-specific abstractions, such as \texttt{DSPy}, requires rewriting pipelines into these specific structures, which entails significant overhead.

\paragraph{Case II: End-to-End LLM Application Development.}
\textit{A practitioner builds an LLM-based system from scratch, jointly designing prompt logic, program structure, and possibly training or fine-tuning stages.} \texttt{promptolution} intentionally only optimizes prompts. In particular, \texttt{DSPy} is better suited to this setting, allowing the entire pipeline to be composed and optimized in a single, abstract program rather than manually engineering and optimizing each step; however, at the cost of coupling the system design to \texttt{DSPy}’s abstractions and losing portability to other frameworks.

\paragraph{Case III: Prompt Optimizer Benchmarking.}
\textit{A researcher wants to systematically benchmark prompt optimization algorithms or develop a new prompt optimizer and perform a comparative evaluation.}
\texttt{promptolution} is advantageous due to its low abstraction level, modular optimizer design, and explicit support for reproducible experiments. Extension to new optimizers, tasks, or evaluation strategies is explicitly encouraged and can be evaluated in a reproducible and controlled manner. Full trajectories, intermediate prompts, and token usage through callbacks and logging enable detailed analysis, setting \texttt{promptolution} apart from other libraries in this regard.

\paragraph{Case IV: Integration in LLM Benchmarking.}
\textit{A researcher conducts systematic benchmarking of LLMs for a new use case. Instead of treating prompts as fixed, they want to include prompt optimization in the benchmark pipeline to account for the strong influence of prompts on performance and reduce variance in findings due to arbitrary prompt choices.}
\texttt{promptolution} allows seamless integration of prompt optimization into any benchmark pipeline, exchanging LLM backends and datasets, and is designed for controlled, systematic, reproducible studies across multiple seeds.

\section{Conclusion \& Future Directions}\label{sec:conclusion}

In this work, we introduced \texttt{promptolution}, a unified and modular open-source Python framework for automatic prompt optimization, designed for both practitioners performing single-task prompt optimization and for researchers conducting systematic experiments. We highlighted the unique position of our framework within the landscape of prompt optimization tools, emphasizing its extensible, modular design and low abstraction level, and its focus on optimizing prompts with non-invasive integration. Furthermore, we demonstrated the role of each component in the framework and how they interact to support effective prompt optimization. Through a comparative evaluation, we verified the utility and competitiveness of \texttt{promptolution}. Finally, we provided concrete use and anti-use cases, highlighting the considerations to take into account when choosing a prompt optimization framework.

Looking ahead, we plan to develop interfaces with higher-level frameworks such as \texttt{DSPy}, enabling a combination of strengths from both ecosystems. To further simplify the management of complex, large-scale experimental setups, we intend to introduce an interface to configuration management frameworks such as \texttt{hydra}~\citep{hydra}. We also aim to improve accessibility by implementing a graphical interface for real-time experiment tracking and visual analysis of the optimization process. Following the recent rise of multi-agent systems, we plan to support the optimization of system prompts and interaction protocols across multiple agents. Additionally, inspired by AutoML, we intend to explore ensembling strategies for prompt optimizers, akin to ELPO~\citep{zhang-arxiv25a}. The extensibility of our framework ensures that we can continue to incorporate new state-of-the-art optimization methods as the field evolves.

\section*{Broader Impact}

By unifying multiple prompt optimization methods that were previously scattered across separate research repositories, \texttt{promptolution} makes the benefits of prompt optimization for improving LLM performance broadly accessible, allowing practitioners to leverage these capabilities in real-world industry applications. At the same time, its modular and extensible design, combined with experiment-friendly implementations, makes the library a powerful tool for researchers benchmarking new prompt optimization algorithms. Since reimplementing competing methods and setting up rigorous benchmark experiments is typically time-consuming, \texttt{promptolution} offers the potential to accelerate research progress in prompt optimization and to enhance methodological comparability across the field.

\section*{Acknowledgments}

Tom Zehle received funding by the European Union. Views and opinions expressed are, however, those of the author(s) only and do not necessarily reflect those of the European Union or the European Commission. Neither the European Union nor the European Commission can be held responsible for them. This work was supported by the European Union’s Horizon Europe research and innovation programm under grant agreement No 101214398 (ELLIOT).

Matthias Aßenmacher received funding from the BERD@NFDI consortium in the context of the work of the National Research Data Infrastructure (NFDI) Association. NFDI is funded by the Federal Republic of Germany and the 16 federal states. The BERD@NFDI consortium is supported within NFDI by the German Research Foundation (DFG) – NFDI 27/1-2026, project number 460037581.

\bibliography{strings,bibliography,proc,my_proc,websites}

\clearpage %
\appendix

\section{Appendix}\label{sec:appendix}

\subsection{Details on Prompt Optimization Tool Categorization}\label{app:categorization}

In the following, we explain the choices (\checkmark/\xmark) in Table~\ref{tab:rw} for all considered open-source frameworks:

\begin{description}[nosep,topsep=5pt]
    \item[DSPy]\citep{khattab-iclr24a}: multiple optimizers (e.g., GEPA, MIPROv2), extensible since adding new optimizers is possible, high abstraction level through declarative programming and monolithic compile method, invasive integration as requiring switching to \texttt{dspy.Module} and \texttt{dspy.Signature} (details in Appendix~\ref{app:dspy}).
    \item[CoolPrompt]\citep{kulin-fruct25a}: multiple optimizers (e.g. DistillPrompt, ReflectivePrompt), partly extensible since implementing own optimizers is possible, but follows no clear framework, high abstraction level as it emphasizes ``zero-configuration'', non-invasive integration since it simply returns a prompt string.
    \item[promptomatix]\citep{murthy-arxiv25a}: uses \texttt{DSPy} optimizers as backend, not extensible by design since it only has one fixed PromptOptimizer class, high abstraction since it builds on top of \texttt{DSPy}, invasive since it wraps \texttt{DSPy}, therefore requiring at least the same effort to integrate.
    \item[prompt-ops]\cite{promptops}: supports a basic and advanced optimizer, not extensible as the advanced optimizer is intended exclusively for Llama models, low abstraction since we can parametrize all components through a single configuration file, non-invasive since it simply returns a prompt string.
    \item[TextGrad]\cite{yuksekgonul2024textgrad}: single optimizer (TextGrad by \citet{yuksekgonul-nature25a}) and thus not extensible, low abstraction as the optimizer is parametrizable in a fine-grained way, invasive integration as requiring prompts to be wrapped in \texttt{textgrad.Variable} and LLM calls to \texttt{textgrad.BlackboxLLM} and engine wrappers.
    \item[AdalFlow]\citep{adalflow}: single optimizer (LLM-AutoDiff by~\citet{yin-arxiv25a}) and thus not extensible, low abstraction as the optimizer is parametrizable in a fine-grained way, invasive integration as it requires re-architecting pipelines into PyTorch-like component classes.
    \item[PromptWizard]\citep{agarwal-arxiv24a}: single optimizer (PromptWizard) and thus not extensible, low abstraction as the optimizer is parametrizable in a fine-grained way, non-invasive since it simply returns a prompt string.
    \item[PromptIM]\citep{promptimizer}: single optimizer (own optimization strategy) and thus not extensible, low abstraction as their optimizer is parametrizable in a fine-grained way, non-invasive since it can either be configured for local experimentation, just returning a prompt string, or to automatically commit the optimized prompt to the LangChain Hub.
\end{description}

\subsection{Delimitation from \texttt{DSPy}}\label{app:dspy}

One might argue that the existing framework \texttt{DSPy}~\citep{khattab-iclr24a} is already very similar to \texttt{promptolution}, as it is also open-source and modular, supports prompt optimization, implements multiple optimizers, supports several LLM implementations, and provides additional utilities. However, there are several important differences, both in underlying purpose and in concrete design, that clearly differentiate the two frameworks:

\begin{description}[nosep,topsep=5pt]
    \item[Different focus]: \texttt{DSPy} includes prompt optimization only as one component within a broader compilation workflow tightly coupled to a program structure, whereas \texttt{promptolution} is not a full application framework and instead focuses exclusively on the prompt optimization stage.
    \item[Abstraction level]: \texttt{DSPy} hides much of the prompt construction behind a high-level declarative interface (though it remains inspectable). In particular, \texttt{DSPy} primarily offers a monolithic \texttt{compile} method that runs the entire optimization routine. In contrast, \texttt{promptolution} provides an iterative \texttt{optimize} routine that repeatedly invokes an optimizer-specific \texttt{\_step} method, along with associated callbacks. This design enables fine-grained control, including intermediate prompt candidates, scores, and token usage. It further facilitates early stopping, custom logging for full transparency, and budget-aware termination through callbacks.
    \item[Target user base]: \texttt{promptolution} is geared toward researchers and advanced ML practitioners, while \texttt{DSPy} primarily targets AI application developers building end-to-end workflows such as agents or RAG systems.
    \item[Experiments]: \texttt{promptolution} makes it easy to implement custom optimizers, tasks, and other components, and is particularly suited for systematic large-scale benchmark experiments due to the integrated config framework. Its extensibility explicitly encourages contributions of new algorithms and tasks, which is not a focus of \texttt{DSPy}.
    \item[Integration \& Portability]: In \texttt{DSPy}, prompt optimization is performed on tasks defined via \texttt{dspy.Module}s (e.g., \texttt{Predict}, \texttt{Refine}, \texttt{ChainOfThought}, \dots) that require \texttt{dspy.Signature}s as input. This couples optimized prompts to \texttt{DSPy}’s program abstraction, which introduces a degree of lock-in when individual subtasks are later migrated to a different system, as additional adaptation may be required. Conversely, \texttt{promptolution} produces a standalone prompt string, allowing optimized prompts to be reused or exchanged across systems with minimal friction, thereby improving portability.
\end{description}

\subsection{Experiment Details}\label{app:exp}

In \S\ref{sec:evaluation}, we evaluate our framework against \texttt{DSPy} and \texttt{AdalFlow}. Since neither provides a straightforward way to restrict compute budgets based on token counts, we enforce token limits by throwing an exception in the respective LLM wrappers when the limit is exceeded, and then returning the last suggested prompt.

For evaluation, we route every LLM API call through the same interface using \texttt{langchain} to ensure a fair comparison of the optimized prompts, and compare exact matches between the predicted and true labels (allowing for differences in capitalization). In the case of \textit{GEPA}, we had to manually clean the LLM outputs because they did not follow the required output format stated in the task description (encapsulating the prediction within <final\_answer> tags). Specifically, we had to remove the ``\texttt{[[ \#\# target \#\# ]]}'' and ``\texttt{[[ \#\# completed \#\# ]]}'' tags. For \texttt{AdalFlow}, we had to intercept API calls due to faulty extraction of system and user prompts. The system-prompt extraction mechanism relied on tags, but these were not forwarded correctly (e.g., ``\texttt{<START\_OF\_USER\_PROMPT>}'' was expected instead of the provided ``\texttt{<START\_OF\_USER>}''). These changes are documented in the experiment repository.
The used LLM was hosted on local servers and accessed via API calls.
The utilized datasets and sample sizes are detailed in Table~\ref{tab:dataset-details}. The dev set is used for optimization, the test set for holdout evaluation of the final prompts. Few-shot examples, if considered by the optimizer, are also taken from the dev set.

\begin{table}[t]
    \centering
    \begin{tabularx}{0.48\textwidth}{XXcc}
        \toprule 
        \textbf{Dataset} & \textbf{Huggingface ID} & $\mathbf{n_\text{dev}}$ & $\mathbf{n_\text{test}}$ \\
        \midrule 
        SST5 & SeetFit/sst5 & 500 & 300 \\
        GSM8K & openai/gsm8k &500  & 300 \\ 
        \bottomrule
    \end{tabularx}
    \caption{Overview of the utilized HuggingFace datasets.}
    \label{tab:dataset-details}
\end{table}

The automatically generated prompts used to compare against a zero-effort baseline are shown in Table~\ref{tab:uno_prompts}. The system prompt accompanying the unoptimized prompts and those optimized by \texttt{promptolution} was ``You are a helpful assistant!''; the system prompts used for the other frameworks were the ones returned after optimization. Note that prompts resulting from \texttt{promptolution} generally do not include any ``\{input\}''-tags (unlike \texttt{DSPy} and \texttt{AdalFlow}, as well as some of the unoptimized prompts) to enable query-independent prompt-caching. Instead, \texttt{promptolution} appends respective queries to the end of the prompt.

The complete experiment code and raw results are publicly available for full reproducibility at \url{https://github.com/finitearth/prompt-optimization-framework-comparison}.

\begin{table}[t!]
    \centering
    \footnotesize
    \setlength{\extrarowheight}{2pt}      %
    \renewcommand{\arraystretch}{1.25}    %
    \begin{tabularx}{0.48\textwidth}{
        @{\hspace{4pt}}c@{\hspace{6pt}}|
        @{\hspace{6pt}}>{\raggedright\arraybackslash}X@{\hspace{4pt}}
    }
        \toprule
        Task & Prompt \\
        \midrule
        \multirow{3}{*}{GSM8K}
        & Solve the maths problem step by step. Give your answer inside
          \texttt{\textless final\_answer\textgreater} \dots
          \texttt{\textless/final\_answer\textgreater}.\textbackslash  n\textbackslash n\texttt{\{input\}} \\
        \cline{2-2}
        & Calculate the solution to the problem below. Show your steps.
        
          \textbackslash  n\textbackslash n\texttt{\{input\}}\textbackslash  n\textbackslash n

          Required output format:
          \texttt{\textless final\_answer\textgreater} ANSWER
          \texttt{\textless/final\_answer\textgreater}. \\
        \cline{2-2}
        & Please analyze the following mathematical problem. Break down your      reasoning into logical steps before stating the solution.
        
          \textbackslash n\textbackslash n Problem: \texttt{\{input\}}\textbackslash n
          
          Format your conclusion as
          \texttt{\textless final\_answer\textgreater}YOUR\_ANSWER
          \texttt{\textless/final\_answer\textgreater}. \\
        \midrule
        \multirow{3}{*}{SST5}
        & Classify the sentiment as very negative, negative, neutral, positive,
          or very positive. Use
          \texttt{\textless final\_answer\textgreater}-tags:
          \textbackslash  n \texttt{\{input\}}\\
        \cline{2-2}
        & Identify the sentiment: very negative, negative, neutral, positive,
          very positive.
          
          \textbackslash n\textbackslash n Text: \texttt{\{input\}}\textbackslash n\textbackslash n
          Answer:
          \texttt{\textless final\_answer\textgreater}label
          \texttt{\textless/final\_answer\textgreater}. \\
        \cline{2-2}
        & Text: \texttt{\{input\}}\textbackslash n\textbackslash n Classify as very negative, negative, neutral, positive, or very positive.
          Output:
          \texttt{\textless final\_answer\textgreater}label
          \texttt{\textless/final\_answer\textgreater}. \\
        \bottomrule
    \end{tabularx}
    \caption{Unoptimized zero-shot prompts per dataset.}
    \label{tab:uno_prompts}
\end{table}

\subsection{Quality Standards}

To ensure high code quality, we adopt established software engineering best practices throughout our package. We maintain a comprehensive test suite that automatically verifies expected behavior after code changes. All tests must pass before a release is published, ensuring users encounter no issues when updating.  
The main branch is protected, and all contributions must be submitted via pull requests, each reviewed by another main contributor. We additionally employ \texttt{pre-commit} hooks to automatically check code formatting, documentation, and other basic quality issues before commits are made, improving readability and maintainability. We also maintain strict documentation standards and do not accept poorly documented pull requests.
Dedicated CI and CI/CD pipelines enable an automated build, test, and release of the package and documentation.

\subsection{Documentation \& Tutorials}

Alongside the open-source software package, we provide extensive documentation available at~\url{https://automl.github.io/promptolution/}. It covers the major components and their functionality, and also includes tutorials that guide users through their first prompt optimization use case. This further enhances the accessibility and usability of our framework.

\subsection{Extensibility of \texttt{promptolution}}

\texttt{promptolution} is designed to be easily extensible, allowing researchers to integrate new optimization algorithms with minimal implementation effort. To illustrate this, we briefly outline how a new prompt optimizer can be added to the framework.

All prompt optimizers inherit from the abstract base class \texttt{BaseOptimizer}, which defines the common interface and execution structure shared across optimization algorithms. Concretely, extending the framework requires implementing only two abstract methods: \texttt{\_pre\_optimization\_loop}, which performs any setup before the optimization begins, and \texttt{\_step}, which executes a single optimization iteration and returns the updated prompt candidate(s). The overall optimization loop, including configuration handling, callback invocation, and other features, is already implemented in the base class via the \texttt{optimize} method.

This design ensures that developers of new optimizers can focus exclusively on their algorithmic logic, while benefiting from shared infrastructure such as logging, callbacks, and budget-aware termination. A similar extension pattern applies when adding new task-, LLM-, or predictor components.

\subsection{System Demonstration}
The system demonstration under \url{https://youtu.be/gySdgjEhsZA} uses the following code example, through which we guide step-by-step. It requires installing the \texttt{promptolution} package with API support enabled. The subsequent imports establish the necessary components for the LLM wrapper, CAPO, and the evaluation task definitions.

\begin{codebox}
! pip install promptolution[api]
\end{codebox}
\label{listing:install}
\vspace{-0.8em}

\begin{codebox}
from promptolution.llms import APILLM
from promptolution.optimizers import CAPO
from promptolution.tasks import JudgeTask
from promptolution.predictors import MarkerBasedPredictor
from promptolution.utils import create_prompts_from_task_description
import pandas as pd
\end{codebox}
\label{listing:imports}

\noindent The dataset, containing raw email-generation instructions, is loaded from a CSV file using \texttt{pandas}:

\begin{codebox}
df_emails = pd.read_csv("emails.csv")
\end{codebox}

\noindent After loading the data, we define a task description that specifies the desired persona, tone, and formatting constraints (including the closing signature). This serves as the reference specification for the entire workflow.

\begin{codebox}
task_description = """Write concise, polite, professional academic emails for me as a PhD student, asking clarifying questions when my instructions are vague, avoiding cliche openings, and always ending with Yours sincerely, Tom Zehle."""
\end{codebox}

\noindent We then instantiate the underlying LLM. In this example, the \texttt{APILLM} wrapper connects to the \texttt{Llama-3.3-70B-Instruct-Turbo} model via the DeepInfra API, which serves as the downstream LLM, the judge, and the meta-LLM.

\begin{codebox}
llm = APILLM(
    model_id="meta-llama/Llama-3.3-70B-"\
             "Instruct-Turbo",
    api_url="https://api.deepinfra.com/"\
            "v1/openai",
    api_key=open("token.txt").read(),
)
\end{codebox}

\noindent To initialize the optimization search space, a set of candidate prompts is derived directly from the task description using the framework's utility function. These serve as the starting population for the optimizer we use later.

\begin{codebox}
initial_prompts = create_prompts_from_task_description(
    task_description,
    llm,
)
\end{codebox}

\noindent Given the subjective nature of the email generation task (lacking a ground truth), a \texttt{JudgeTask} is configured. This setup utilizes an LLM-as-a-Judge approach to evaluate the semantic quality of the generated outputs against the input instructions. An alternative could be the \texttt{ClassificationTask} when ground truth labels are available, or the \texttt{RewardTask} when the user can define an objective function to score the outputs. 

\begin{codebox}
task = JudgeTask(
    df_emails,
    judge_llm=llm,
    task_description=task_description,
    x_column="instruction",
)
\end{codebox}

\noindent Next, we instantiate the \texttt{CAPO} optimizer with a marker-based predictor. The optimization routine is executed for six iterations (\texttt{n\_steps=6}) to refine the prompt candidates iteratively.

\begin{codebox}
optimizer = CAPO(
    task=task,
    predictor=MarkerBasedPredictor(llm),
    meta_llm=llm,
    initial_prompts=initial_prompts,
    check_fs_accuracy=False
)
final_prompts = optimizer.optimize(n_steps=6)
\end{codebox}

\noindent After optimization completes, the final optimized prompts are returned and ready for use in email generation.

\end{document}